\documentclass[]{llncs}
\pdfoutput=1
\usepackage{graphicx}
\usepackage{amsmath,amssymb} 
\usepackage{color}
\usepackage[width=122mm,left=12mm,paperwidth=146mm,height=193mm,top=12mm,paperheight=217mm]{geometry}

\title{Large Field and High Resolution: Detecting Needle in Haystack}

\author{ \large Hadar Gorodissky \and Daniel Harari \and Shimon Ullman }

\institute{Department of Computer Science and Applied Mathematics, \\ Weizmann Institute of Science, Rehovot, Israel \\ gorohadar@gmail.com, \{hararid,shimon.ullman\}@weizmann.ac.il 
}

\begin{document}
\pagestyle{headings}
\mainmatter

\maketitle

\begin{abstract}
The growing use of convolutional neural networks (CNN) for a broad range of visual tasks, including tasks involving fine details, raises the problem of applying such networks to a large field of view, since the amount of computations increases significantly with the number of pixels.  To deal effectively with this difficulty, we develop and compare methods of using CNNs for the task of small target localization in natural images, given a limited "budget" of samples to form an image. Inspired in part by human vision, we develop and compare variable sampling schemes, with peak resolution at the center and decreasing resolution with eccentricity, applied iteratively by re-centering the image at the previous predicted target location. The results indicate that variable resolution models significantly outperform constant resolution models. Surprisingly, variable resolution models and in particular multi-channel models, outperform the optimal, "budget-free" full-resolution model, using only 5\% of the samples.


\keywords{variable resolution, target localization, multi-resolution, image sampling, large field of view, fine details, human vision}
\end{abstract}

\section{Introduction}
\label{sect:intro}
A general, flexible visual system faces the dilemma of combining high-resolution with a large field of view (FOV). For example, to notice an approaching car, which might pose a danger, the visual FOV should be as large as possible. On the other hand, to guide the task of passing a thread through the eye of a needle, the visual system requires exquisite resolution. In the human visual system, the FOV spans about 120 degrees, with peak resolution of about 0.5 minutes of arc~\cite{barton2003,howard1995}. Such opposing requirements are difficult to reconcile: covering the field at the highest resolution will require the visual system to operate on images with a size of over 200 million image sampling points. For current machine vision systems, which process images at a typical size around ${500\times500}$ pixels by deep convolutional neural network (CNN) models, this will require a scaling of the processing power by three orders of magnitude, since the amount of computations often scales about linearly with the number of samplings \cite{Mnih2014}. For biological systems, such as the human visual system, this requirement is anatomically impossible, requiring for example the diameter of the optic nerve (carrying information from the eye to the brain) to increase in diameter from 3-5mm to almost 10cm \cite{Sylvester1961}. Evolution approached this resolution-size dilemma by: (1) acquiring images at variable resolutions, with high resolution in a small region at the center of the visual field, and reduced resolution as a function of eccentricity (distance from the center of the visual field), and (2) the use of eye movements, i.e. re-centering of the visual field around different fixation points.

\begin{figure}[t]
\centering
\includegraphics[width=0.7\columnwidth]{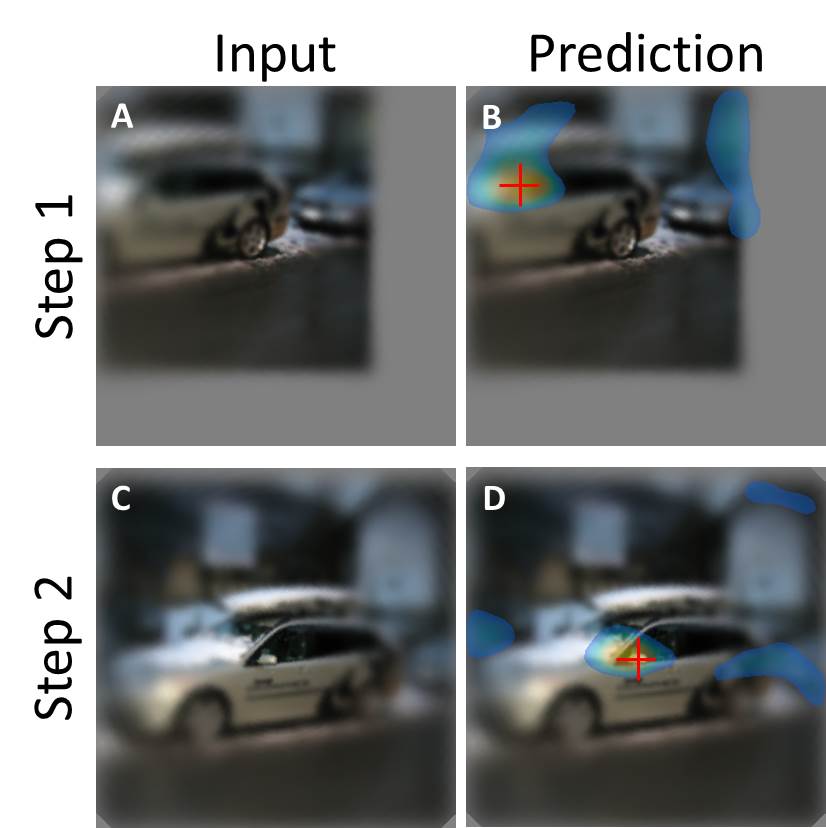}
\caption{\textbf{Iterative variable resolution scheme with image re-centering.} Example of the first and second processing steps of our scheme for locating a car's side-mirror (demonstrated for variable single channel model). \textbf{(A)} Initial input image at the first step, \textbf{(B)} the model's probability map (last deconvolution layer) and target prediction after the first step, \textbf{(C)} the successive step input image is centered around the target's predicted location from the first step, \textbf{(D)} the model's probability map and target prediction after the second step.}
\label{fig:saccade}
\end{figure}

In this work, we address the problem of obtaining both a large field of view as well as high resolution, under the limitation of a fixed pixel "budget". We focus on the task of detecting a small target in natural images, which requires the ability to process a large field of view as well as fine details at high resolution. We will assume that the system uses a total number of K sampling points to form an image, but can distribute the sampling points in different configurations, in which the sampling resolution is not necessarily constant across the visual field. We compare different designs of distribution of the sampling points across the visual field. The common approach in computer vision is to use constant sampling, namely images with uniform regular pixel grid, which spreads the pixels budget uniformly across the desired field of view. Such constant resolution images produce a simple global trade-off between field of view and resolution: using a fixed number of sampling points, can produce a low resolution image with a large field of view, or a high resolution image with a proportionally smaller field of view, hence requiring the acquisition of multiple such high-resolution images to cover the desired field of view, while losing context-related information of the whole field of view. 

We develop for this purpose an approach, inspired in part by the human visual system, which utilizes a non-uniform acquisition of the visual field in a sequence of directed fixations. Our approach makes the most of the limited available resources in solving the visual task of small target detection, by making use of variable sampling, i.e. processing images with peak resolution at the center, which decreases with eccentricity, and re-centering iterations, similar to eye movements (Fig. \ref{fig:saccade}). We further compare the use of a single variable channel with multiple uniform sampling channels, showing a superiority of the multi-channel approach.  

\section{Previous work}
\label{sect:prev_work}

Early studies used variable resolution processing, with high resolution at the center and decreasing resolution towards the periphery, to solve classical computer vision problems such as the correspondence problem for shape matching. Berg and Malik~\cite{Berg2005} introduced the \textit{geometric blur} filter and used it for computing points correspondence. The geometric blur filter is a smoothed version of the signal around a feature point, blurred by a spatially varying kernel with increasingly larger blur away from that point. Features based on the geometric blur descriptor yielded impressive improvement in object recognition with deformable shape matching capabilities \cite{Berg2005}.

In the past decade, researchers have addressed the problem of the high computational cost in applying various computer vision schemes, and in particular  neural networks, to large images at full resolution. Several studies suggested models, inspired by the human non-uniform image acquisition in the retina and saccadic eye movements, which improve the invariance to scale, translation and clutter in classification tasks, with a significantly reduced computational resources~\cite{Larochelle2010,Mnih2014}. 
Larochelle and Hinton \cite{Larochelle2010} proposed a Boltzmann machine with third-order connections that can learn how to accumulate information about a shape over several fixations. Their model used a retinal window, which only had enough high resolution pixels to cover a small area of the image. The model then combines information from multiple glimpses of the same object. The resulting capabilities were illustrated on example tasks of digit classification and facial expressions in small images.
Mnih et al. \cite{Mnih2014} used a visual attention model formulated as a single recurrent neural network, which takes a glimpse window around a center location (concatenated images at decreasing resolutions and increasing fields of views) as its input. The model then uses the internal state of the network to select the location and input parameters of the next glimpse.  The system was trained end-to-end from pixel inputs to actions  and demonstrated on a cluttered digit classification task in images from the MNIST dataset.

Networks efficiency and performance can also be improved by introducing task-specific inductive biases to the processing scheme. Recent studies on the expressive efficiency and inductive bias of various architectural features in CNNs, including depth, width and pooling geometry, show that the architecture of the network highly affects its ability to succeed in a given task \cite{Cohen2016a,Cohen2016b}. In the final discussion, we suggest connections between these results and the current study, in particular, regarding the surprising efficiency of variable resolution methods compared with a baseline of unlimited full resolution model.

The variable resolution models used in this work are inspired by models of the sampling mechanisms in the human visual system \cite{Wilson1979,Poggio2014}. 
Poggio et al. \cite{Poggio2014} modeled a linear dependency between eccentricity and the receptive field size of retinal  ganglion cells, as part of a sampling extension of a visual invariance theory. The so-called \textit{truncated inverted pyramid} model, consists of cells with fixed receptive field size near the center of fixation, and cells with linearly increasing receptive fields as a function of eccentricity outside the center region. Using this model, we can estimate the number of available sampling points and their spatial configuration for a given FOV. An earlier study by Wilson et al. \cite{Wilson1979} proposed a model for patterns detection threshold based upon probability summation of several filter channels with different sizes and resolutions. An analysis of human psychophysics supported the existence of 4-5 parallel size-tuned mechanisms centered at each point, which increase in size linearly with eccentricity. 

In this work, we focus on deploying CNNs to images with large FOV combined  with high resolution, given a limited 'budget' of image samples. We compare between two main approaches, variable versus constant resolution models, for distributing the pixel budget. We evaluate the performance of the models on a natural task of finding small target of interest in images.

\section{Methods}
\label{sect:methods}

\subsection{Overview}
\label{sect:overview}

We study models with different input configurations, given a fixed budget of image samples and with the same amount of computation resources. In particular, we compare input configurations of constant versus variable resolution. The constant resolution input configuration uses a uniform regular grid across the desired FOV for sampling the input image with the given pixel budget. The variable resolution alternative uses a non-uniform grid, with peak resolution at the center of the desired FOV. The input sampling configurations are further described in section \ref{sect:samp_conf}.
We focus on the task of finding small target of interest since the task naturally combines the ability to process fine details at high resolution with the ability to also process a large FOV, where the target can be located. Section \ref{sect:task_descr} provides more details on this visual task.
To provide the same amount of resources across models, we base all models on the same network architecture. In particular, we adopt a single CNN to various input configurations using the common practice in transfer learning \cite{goodfellow2016} and a common training procedure. The trained models are applied iteratively until convergence, by re-centering the input's FOV on the predicted target location of the previous step. The models' architecture as well as training and testing schemes are described in section \ref{sect:model}.

\begin{figure}[t]
\centering
\includegraphics[width=1\columnwidth]{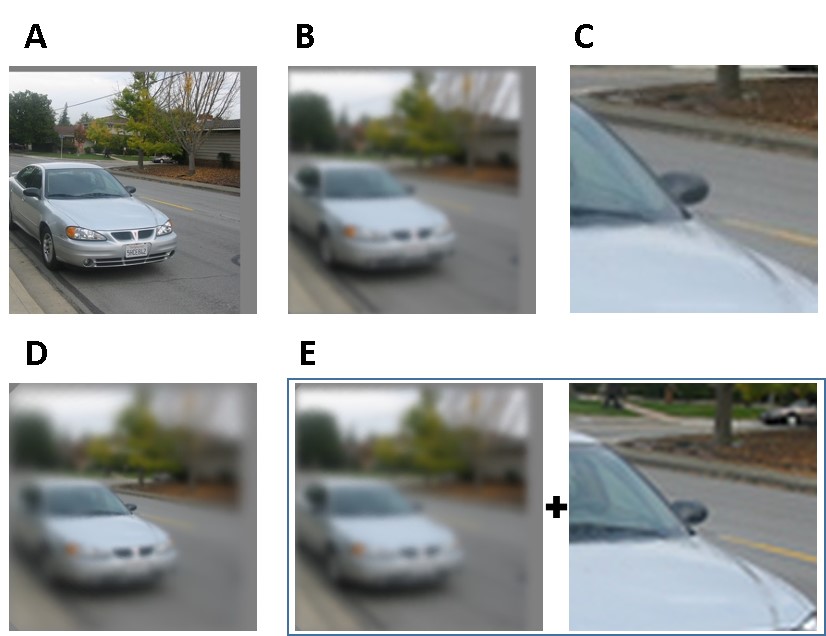}
\caption{\textbf{Variable vs. constant resolution.} Examples of input  images, sampled for the following models: \textbf{(A)} "Budget free" full (original) resolution model, used as a baseline; \textbf{(B)} Constant resolution with large FOV; \textbf{(C)} Constant resolution with small FOV; \textbf{(D)} Variable resolution with a single channel; \textbf{(E)} Variable resolution with multiple channels. The models in \textbf{(B-E)} use a fixed budget with the same number of sampling points. All images are interpolated to a common size, regardless of the sampling configuration.}
\label{fig:Input example}
\end{figure}

\subsection{Input sampling configurations}
\label{sect:samp_conf}

We consider several sampling configurations of the input images given a fixed budget of samples, and train a different model for each of the input sampling configurations. The original images in our augmented dataset are of size \(384\times384\) samples. We refer to the given resolution of the original images as 'full resolution', and to the FOV spanned by the original image pixels, as the 'full FOV' of scenes depicted in these images (Fig. \ref{fig:Input example}A). To determine a realistic sample budget for our scheme, we used the variable resolution human model proposed by Poggio et al. \cite{Poggio2014}. The model yields a budget of 7630 samples within the full FOV. 

The common sampling alternative in current schemes is constant resolution, where the budget of samples is uniformly distributed on a regular grid. To cover a large FOV with a limited budget of samples (7630), the sampling distance must be large, resulting in low resolution images. On the other hand, to gain the highest resolution, the samples should be densely distributed, resulting in a small FOV. In our settings, the \textbf{large FOV constant model} is trained on images covering the full FOV at the cost of a reduced resolution, with a sampling distance of 4.4 pixels (Fig. \ref{fig:Input example}B). The \textbf{small FOV constant model} is trained on images sampled densely at the highest possible resolution within a small FOV around the center ($\sqrt{7630}\cong87$ pixels wide, Fig. \ref{fig:Input example}C).

An alternative to the common sampling approach is variable resolution, where the budget of samples is distributed in a non-uniform manner across the desired FOV. The variable resolution approach allows the combination of high resolution with a large FOV. A visual system equipped with such sampling capabilities may utilize a single sampling mechanism or multiple sampling mechanisms operating in parallel \cite{Wilson1979}. In our settings, the \textbf{single channel model} (Fig. \ref{fig:Input example}D) is trained on images sampled densely around the center and with continuously decreasing density as a function of eccentricity (i.e. distance from the center) \cite{Poggio2014}. The \textbf{multi-channel model} is trained on images sampled at multiple sampling rates to form multiple input channels, where each channel has a different uniform sampling rate. In this model, the input channels cover co-centered image regions with increasing FOVs, while sharing the given pixel budget. We demonstrate the capabilities of this model in our experiments using 2 channels: one channel, which is densely sampled using \(60\times60\) samples and spans a small FOV (\(120\times120\)), and another channel, which spans the full FOV (\(384\times384\)) using a coarse resolution with a sampling distance of 6.4 pixels (Fig. \ref{fig:Input example}E). The number of samples used by the two channels together sums up to the given pixel budget. Samples of coarse channels, which fall within the region of the smaller channel, are obtained by interpolation from the higher resolution samples.

We use a reference model (\textbf{full resolution model}) as a baseline. The baseline model has the same architecture and training procedure as the other models above. However, unlike the other models, we remove the pixel budget limitation on the input to obtain the finest resolution across the full FOV, using  \(20\) times more samples (Fig. \ref{fig:Input example}A).

\subsection{Visual target localization task}
\label{sect:task_descr}

We focus on the task of visual target localization, in which the goal is to identify and locate instances of a predefined target of interest in images. In our experiments, we specifically use small object parts occupying less than 1\% of the input image. We found this task interesting for the following reasons: (1) Target localization is a fundamental visual task in many real life scenarios, (2) recognition and localization of the target object part depend on its relations to other parts of the object (i.e. the part on its own may be hard to recognize and locate), (3) this task requires the integration of both coarse contextual information of the target's surroundings as well as fine details of the target appearance. Object parts are relatively small compared with the whole object and often exhibit high variability in appearance. As a result, these parts often cannot be identified and processed as independent objects for detection, classification, and segmentation.

\begin{figure}[t]
\centering
\includegraphics[width=1\columnwidth]{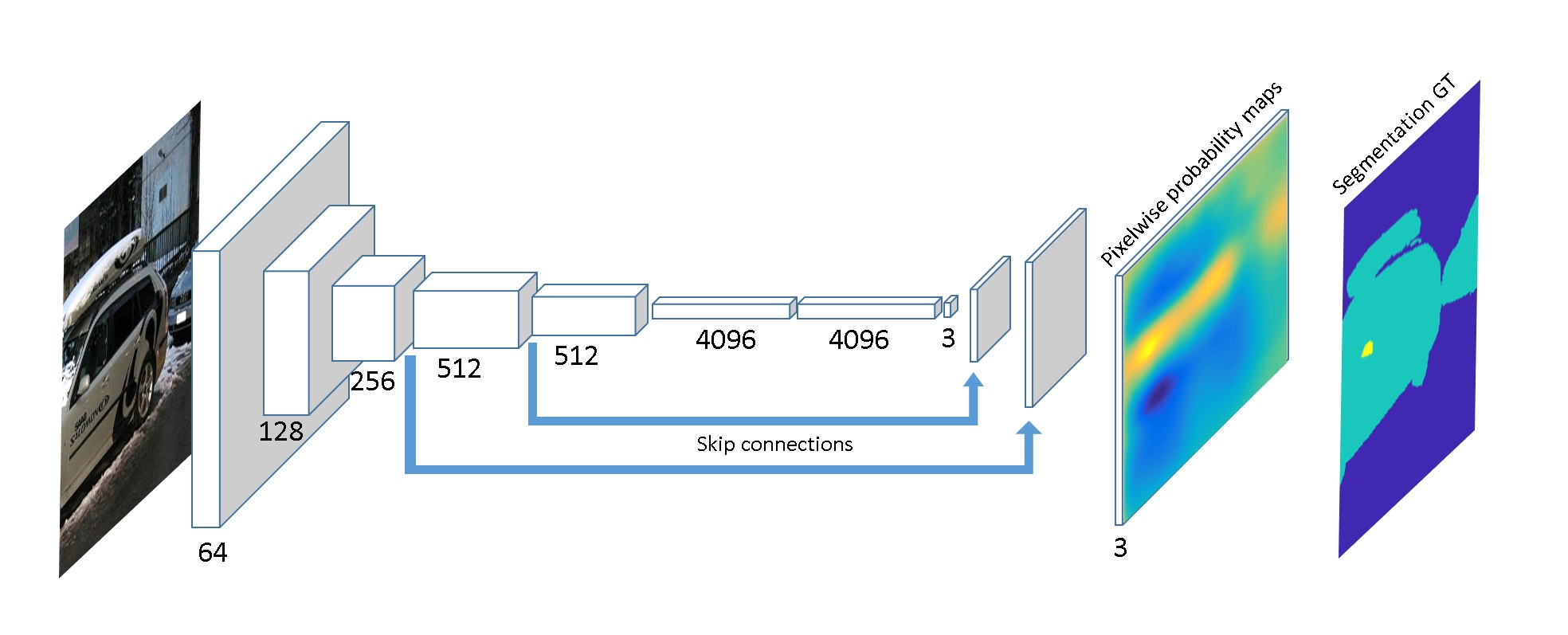}
\caption{\textbf{Model architecture.} We adopt the full convolutional neural network (FCN) model for semantic segmentation. The network is based on the VGG-16 classification network \cite{Simonyan2014} and extended by three deconvolution blocks with skip connections as described in \cite{Shelhamer2017}. In our settings, we use three class labels for the target part, object containing the part and background (a car side-mirror target is shown).}
\label{fig:model_arch}
\end{figure}

\subsection{Architecture and processing scheme}
\label{sect:model}

We adopt a CNN model by Shelhamer et al. \cite{Shelhamer2017} for semantic segmentation (a.k.a. FCN, fully convolutional network). The model, which is broadly used as a basic model for the task of part segmentation \cite{Wang2015,Xia2016,Chen2015}, infers the class of each pixel in the input image from a predefined list of semantic class labels. The network is based on a pre-trained VGG-16 classification network \cite{Simonyan2014}, which is extended to the FCN-8s segmentation network, as described in \cite{Shelhamer2017} and shown in figure \ref{fig:model_arch}. We found this model to be competitive with other alternative architectures for our task of target detection. 

For training, we apply the common practice in transfer learning \cite{goodfellow2016}, by fine-tuning the parameters of the last 3 \textit{convolutional} layers and the \textit{skip} layers, while keeping the parameters of the other layers fixed. We train different models for each of the input sampling configurations as described in section \ref{sect:samp_conf}. To allow efficient utilization of all network resources for the various FOVs, we interpolate the input images to a common size of \(384\times384\) for all models, regardless of the sampling configuration. Interestingly, a similar scaling, termed cortical magnification, is used in the human and primate visual systems \cite{Poggio2014,Cowey1974,Hubel1974}. Training annotations consist of three class labels, including the target part, the object containing the target part and the background. The classes we considered are highly unbalanced, since on average, the target of interest occupies only about 0.27\% of the image size. We therefore used a weighted loss for optimization, which works better than regular cross entropy, by allocating the same total weights to all classes, and uniformly distributing the weights of each class across all the class pixels.

The target position is predicted at the spatial location of the peak in the probability map of the target of interest class in the last layer of the network. The multi-channel model (section \ref{sect:samp_conf}), computes a probability map for each channel (using the corresponding input sampling), and the target position is predicted at the peak spatial location of the  probability map across all channels. All models are applied iteratively, where at each iteration a new input image is produced by re-centering the FOV around the predicted target location of the previous iteration (Fig. \ref{fig:saccade}, \ref{fig:mc_demo}). 

To evaluate the predicted target location, we measure the Euclidean distance between the spatial coordinates of the prediction point and the closest point in the target's annotation mask. 

\section{Experimental evaluation}
\label{sect:exp_eval}

\subsection{Data and Augmentation}
\label{sect:data}

In our experiments, we used images from the PASCAL-Part dataset \cite{Chen2014} with the following targets of interest: car's side-mirror, car's headlights and cat's ear. For each target of interest we had approximately 350 images containing roughly 500 targets. We used 70\% of the images for training and 30\% for testing. To have a sufficient number of training examples, we augmented the dataset, by cropping sub-image regions of size \(384\times384\) at random locations from the original images, which are of average size of \(367\times486\) (plain padding is used whenever the cropping exceeds the original image FOV). Each augmented image included all the pixels of the target part and at least 60\% of the object containing the part, as seen in the original image. Overall, the augmented dataset, consisted of about 12,000 training and testing examples for each of our 3 targets of interest.
For each model type, both training and testing input images are first sampled according to the model's sampling configuration, as described in section \ref{sect:samp_conf}. 

\subsection{Detailed results analysis}
\label{sect:results}
The following two subsections present a detailed analysis of the results related to the target category of car's side-mirror. Additional results are presented in section \ref{sect:other_targets} for two other target categories: car's headlights and cat's ear.

\subsubsection{First Step Results.}
In our scheme, models are initially applied to images with random target positions from our augmented dataset (section \ref{sect:data}). We refer to this processing phase as the 'first step' in the iterative scheme. We define the 'initial error' for a given input image, to be the distance in pixels between the center coordinates of the image and the closest target point. The prediction error is defined as the distance in pixels between the predicted and true target locations as described in section \ref{sect:model}. 

\begin{figure}[t]
\centering
\includegraphics[width=1\columnwidth]{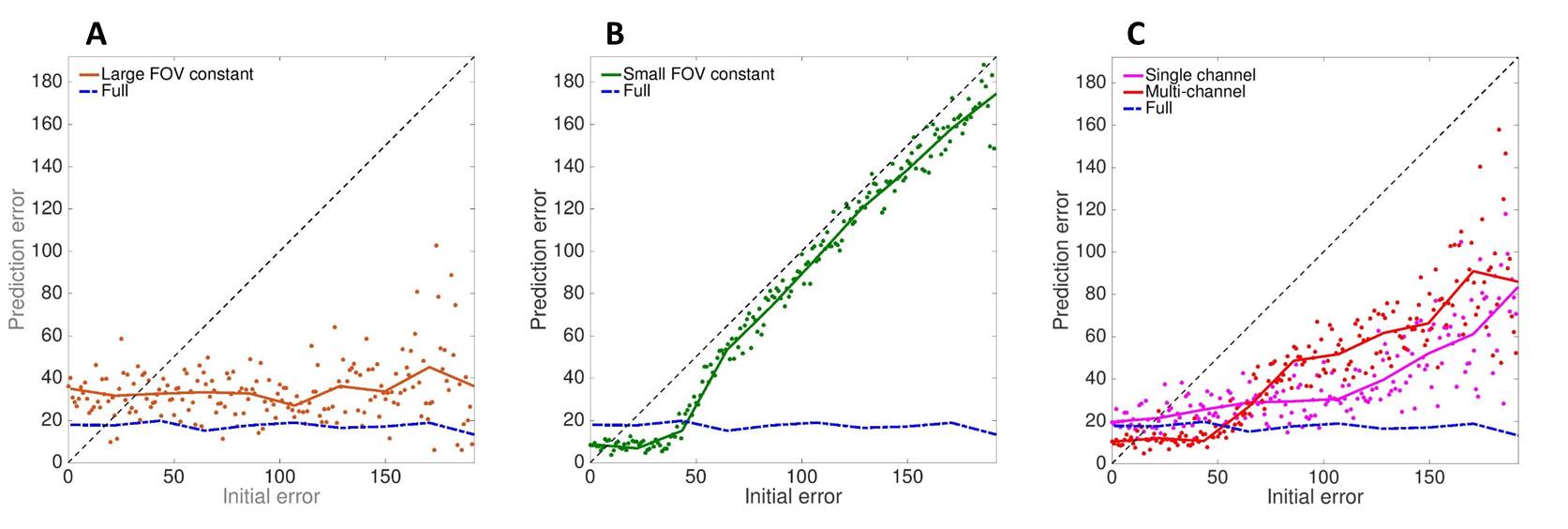}
\caption{\textbf{First step prediction error analysis.} Mean prediction error (distance in pixels between predicted and true target locations; vertical axis), against the initial target error (distance in pixels between the true target location and the image center; horizontal axis). \textbf{(A)} The large FOV constant resolution model vs. the baseline-full model. \textbf{(B)} The small FOV constant resolution model vs. the baseline-full model. \textbf{(C)} The single and multi-channel variable resolution models vs. the baseline-full model. The scatter plots present the mean prediction error of instances with initial error in the range of one pixel.  The curves present a piecewise linear fit (10 pieces) of the mean prediction error.}
\label{fig:step1_results}
\end{figure}

Figure \ref{fig:step1_results} presents the prediction error versus the initial error of the constant and variable resolution models, as well as the baseline \textbf{full-resolution} model. The baseline model is "budget-free", while the sample resources of other models are limited by the given budget and use only about 5\% of the samples compared with the baseline (see section \ref{sect:samp_conf}). The results show that both the \textbf{large FOV constant} model and \textbf{full resolution} baseline model, are roughly invariant to the target location in the image, and hence yield practically the same prediction error for any target location in the image, i.e. any initial error (Fig. \ref{fig:step1_results}A). This result is expected due to the fully convolutional architecture of the underlying network and the nature of the constant resolution input images, where the samples are distributed uniformly across the FOV. The results also indicate that the available sample resources affect the model performance, as demonstrated by the superior performance of the baseline model compared with the \textbf{large FOV constant} model. 

Figure \ref{fig:step1_results}B shows that the alternative \textbf{small FOV constant} model outperforms both the \textbf{large FOV constant} as well as the baseline models, but only when the target falls within the small FOV (initial error of 40 pixels or less). Otherwise, which is the majority of the cases, this model fails to improve on the initial error.

Figure \ref{fig:step1_results}C shows the performance of the two variable resolution models compared with the baseline model. The results show that both the single and multi-channel models are sensitive to the target distance from the image center, and the predictions for images with targets close to the image center are significantly better than for images with random target locations (\(t(399)=7.686, p<10^{-3})\). Furthermore, the results show that the \textbf{single channel variable} model converges to the same error as the baseline model when the initial error is small,  and that the \textbf{multi-channel variable} model outperforms even the baseline model when the initial error is small. 

We conclude that at this 'first step', given the limited sample budget, none of the models has superior performance across the entire range of initial errors over the other models. However, while the predictions of the constant resolution models are independent of the initial error, the predictions of the variable resolution models improve as the initial error decreases and the target gets closer to the image center. This ability suggests that an iterative saccadic mechanism, which utilizes the variable resolution models in successive steps, can improve the system performance drastically. We validate this capability in the next subsection.

\begin{figure}[t]
\centering
\includegraphics[width=1\columnwidth]{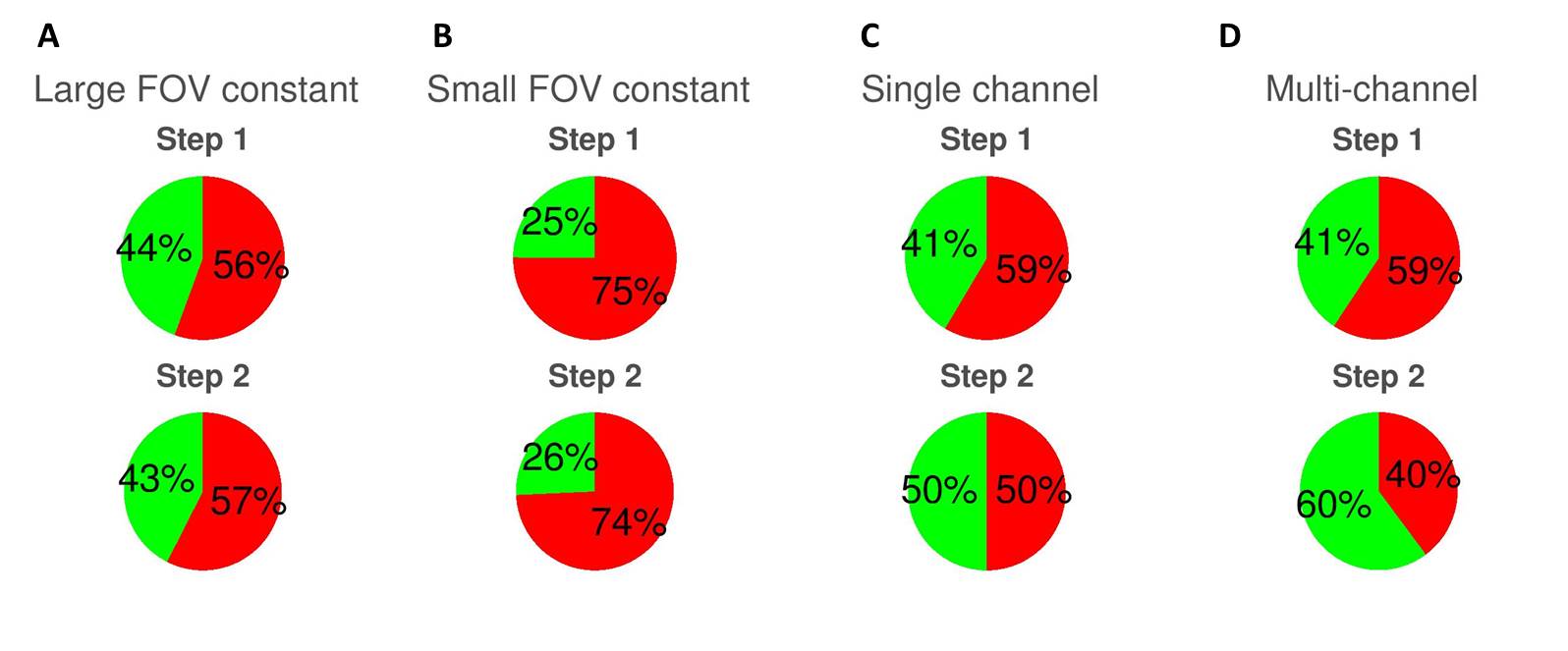}
\caption{\textbf{Models vs. baseline accuracy.} Pie charts representing models accuracy compared with the baseline-full resolution model, after the first and second processing steps. \textbf{(A)} large FOV constant model; \textbf{(B)} small FOV constant model; \textbf{(C)} single channel variable model; \textbf{(D)} multi-channel variable model. The green part represents the fraction of trials, for which the models achieved a better accuracy than the baseline, and the red part represents the fraction of trials, for which the models achieved a degraded accuracy compared with the baseline-full model. The data for these charts consist of a subset of 5200 images from our test set, with equal number of images across all initial error values.}
\label{fig:step2_results}
\end{figure}

\subsubsection{Additional step}
Following the 'first step' described in the previous subsection, our scheme iterates the process, by re-applying each model to a translated version of the initial input image, where the FOV is re-centered around the predicted target location from the previous step. The corresponding sampling configuration is applied to the re-centered images at the input of each model. This successive processing on re-centered images, is similar to the human sequential eye movements mechanism.

Figure~\ref{fig:step2_results} compares between the prediction error  of the different models against the baseline model, after the first and second steps. The pie-charts show in green the percentage of images, where models outperform the baseline, and in red the remaining percentage of images, where the baseline has superior accuracy. 

The results show that the constant resolution models do not improve  with additional steps (Fig. \ref{fig:step2_results}A-B). The \textbf{large FOV constant} model "saturates" as a result of its invariance to the target location (initial error). The \textbf{small FOV constant} model lack of improvement is due to its inability to improve over large initial errors, while performing well only within for a short range of small initial errors.  

In contrast with the constant resolution models, the accuracy of the variable models increases with iterations (Fig. \ref{fig:step2_results}C-D). The results show that the second step allows the variables models to outperform the constant models with equal resources. The \textbf{single channel} model increases its accuracy by 25\% in the second step relative to the first step, while the \textbf{multi-channel} improves its accuracy by 50\%. With the second step the \textbf{single channel} model is comparable to the baseline \textbf{full resolution} model, and the \textbf{multi-channel} model outperforms the baseline model in 60\% of the cases. Both variable models use only 5\% of the baseline sample resource. 

Our experiments indicate that the accuracy of the variable models converges in our setting after an average of 1.5 prediction steps, meaning that applying a third step does not contribute significant improvements to the predictions accuracy. 

\begin{figure}[t]
\centering
\includegraphics[width=1\columnwidth]{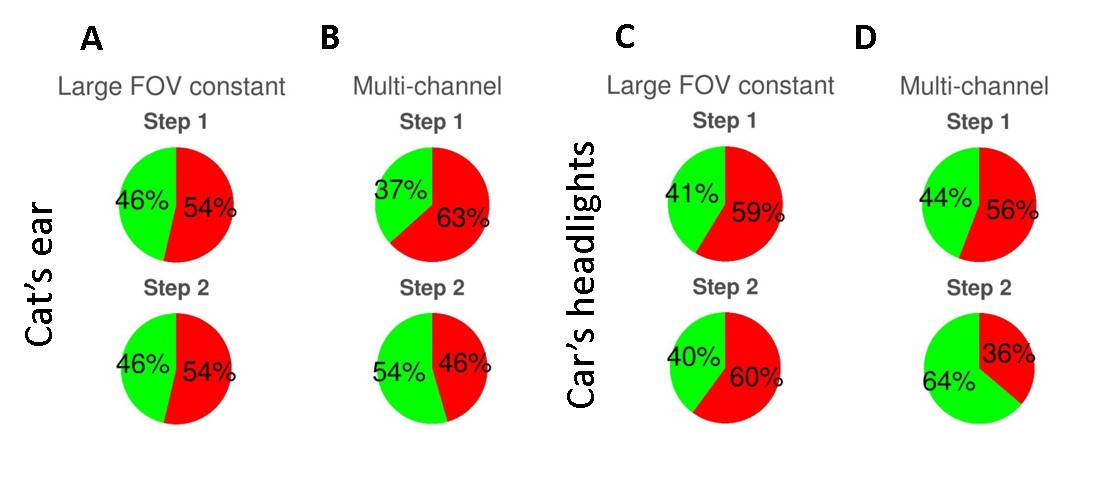}
\caption{\textbf{Models vs. baseline accuracy - other targets.} Pie charts representing models accuracy compared with the baseline-full resolution model, after the first and second processing steps, for two additional target categories: \textbf{(A-B)} cat's ear; \textbf{(C-D)} car's headlights. The analysis is presented only for the large FOV constant resolution model \textbf{(A,C)}, and the multi-channel variable resolution model \textbf{(B,D)}. Chart colors are similar to Figure \ref{fig:step2_results}.}
\label{fig:other_targets}
\end{figure} 

\subsection{Results for other targets}
\label{sect:other_targets}
\begin{figure}
\centering
\includegraphics[scale=0.58]{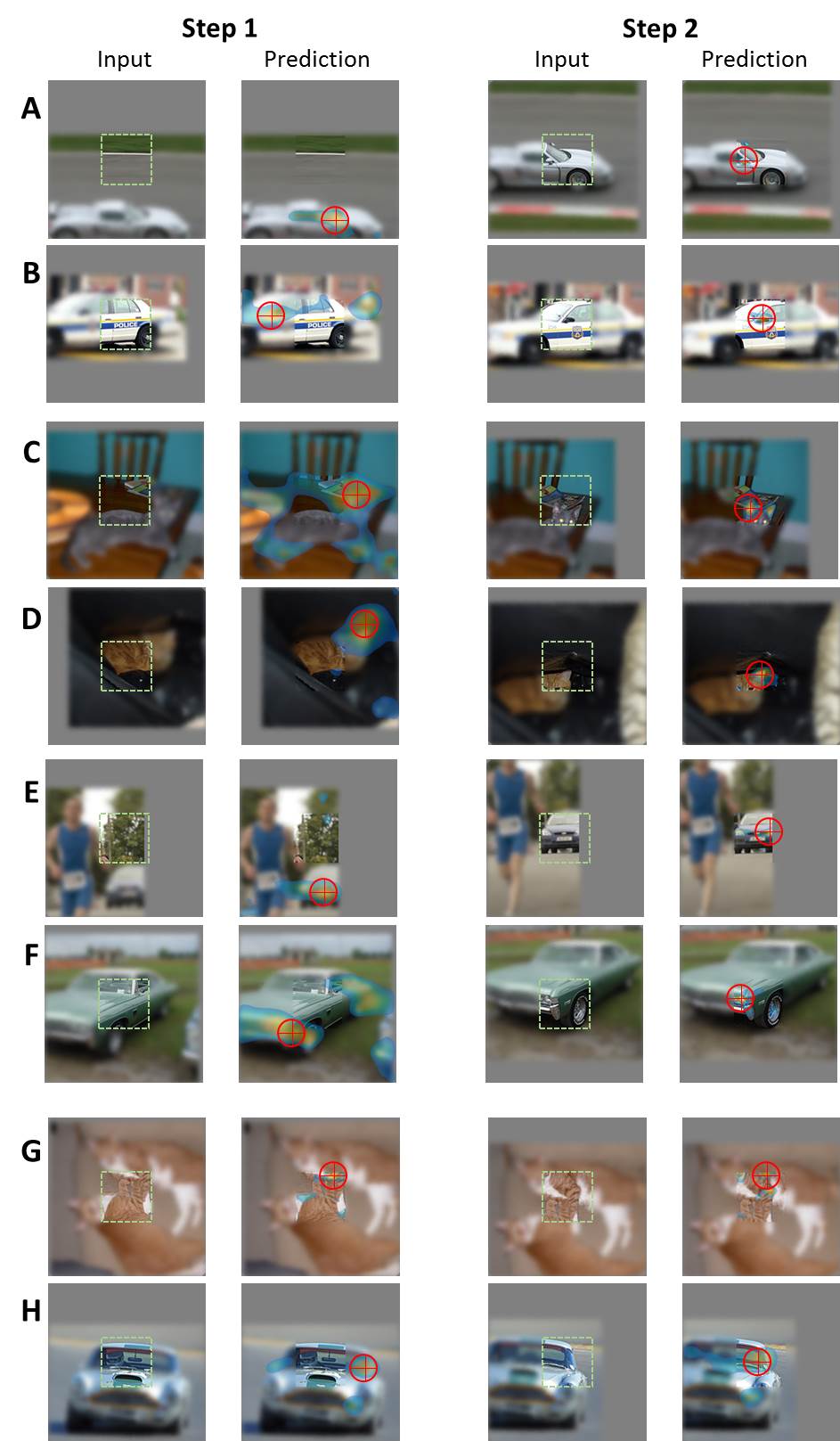}
\caption{\textbf{Multi-channel target prediction examples.} Examples of input images and target predictions at the first and second processing steps of our scheme, by the multi-channel variable resolution model. \textbf{(A-F)} Positive examples of correct target predictions for car's side-mirror \textbf{(A,B)}, cat's ear \textbf{(C,D)} and car's headlights \textbf{(E,F)}. \textbf{(G-H)} False detection examples of cat's ear \textbf{(G)} and car's side-mirror \textbf{(H)}. All input images present the input to the channel with high-resolution and small FOV (marked with a broken-line box) on top of the input to the channel with low-resolution and large FOV.}
\label{fig:mc_demo}
\end{figure}

In this section we test the generality of the results by repeating the experiments described in section \ref{sect:results} for two additional target categories, car's headlights and cat's ear. We use the same underlying network architecture and training procedures used for the car's side-mirror category, to train constant and variable resolution models for the new target categories. Figure \ref{fig:other_targets} presents a comparison between the prediction error of the \textbf{large FOV constant} resolution model and the \textbf{multi-channel} variable resolution model against the baseline \textbf{full resolution} model, after the first and second steps. The results indicate that the models exhibit similar performance characteristics to the car's side-mirror target. Here too, the constant model does not improve after the first step, as opposed to the multi-channel model, which succeeds in outperforming the baseline model after the second step. Figure \ref{fig:mc_demo} presents a few positive examples of correct target localization and a few false alarms, predicted by the multi-channel model applied to the three target categories.

\section{Conclusions}
\label{sect:conclusions}

In this work, we studied the problem of obtaining both large FOV and high resolution at the same time, given a limited ‘budget’ of sampling points to form an image. We compared between models of constant resolution versus models of variable resolution, inspired by human vision. For the constant resolution models, we considered the tradeoff between size and resolution of the sampled image, in particular, covering the entire FOV at low resolution, compared with a limited FOV sampled at a high resolution. For the variable resolution approach, we considered two types of models: a single channel with variable resolution, or multiple channels at constant different resolutions. We focused on the task of localizing small targets of interest, namely, detecting a small object part in a natural image. We chose this task because it naturally combines the requirements of high-resolution, to identify the small target, and large FOV, to detect the target at all possible locations. In developing and testing the target detection models, we employed state-of-the-art deep nets combined with successive fixations, for each configuration type.

Our main results are: (1) variable resolution models achieved better accuracy than the equally resourced constant resolution models. (2) Surprisingly, the variable models achieved the same and even better accuracy level as the baseline-full model, while using only about 5\% of the full model sampling resources. (3) The multi-channel model performed better than the single channel variable model. (4) The variable resolution models required re-centering, or eye movements. Within the tested conditions, an average of 1.5 prediction steps where required (but the number may increase with image size and number of channels).  

The first result demonstrates the advantage of non-uniform sampling over uniform sampling. The variable resolution models appear to use the low-resolution information in the periphery effectively to guide the high-resolution stage, which then predicts the target location accurately. The second result implies that in addition to the allocation of sampling points, there is another factor that contributes to the variable models accuracy, since even with less samples, the variable models achieve the same or better accuracy compared with the full model. We suggest that the additional factor may be involved with a better allocation of the network resources. In networks trained on full resolution images, the resources of the network, measured for example in terms of number of weights allocated to a unit area in the image, are uniformly distributed. In comparison, in the two-channel model, the resources are divided about equally between the two channels, and consequently about half of the resources are allocated to the smaller channel, which covers about 10\% of the full FOV image area. As suggested by recent work,~\cite{Cohen2015,Levine2017} the allocation of network resources can have a significant effect on its performance, and it is conceivable that allocating sufficient resources to the crucial target area contributes to improved performance.  Finally, the multi-channel model outperforms the single channel variable model, possibly due to two reasons. First, unlike variable resolution channel, the mutli-channel scheme uses uniform resolution images, which are easier to train with convolutional neural networks. Second, similar to the discussion above, the mutli-channel scheme may use a better allocation of the network resources compared with the single-channel model.

In future work, it will be interesting to extend the current study is several directions. First, compare different resolution models applied to other visual tasks. Second, develop a more general model of using multiple resolution images by deep nets. Finally, obtain a better theoretical understanding of allocating network resources to different resolution channels.

\section*{Acknowledgements}
This work is supported by Israeli Science Foundation (ISF) grant 320/16 and the German Research Foundation (DFG Grant ZO 349/1-1).

\bibliographystyle{splncs}
\bibliography{Largefield_arxiv}
\end{document}